\documentclass[final,3p,times,twocolumn, sort&compress]{elsarticle}

\usepackage{setspace}
\usepackage{algorithm}
\usepackage{algorithmicx}
\usepackage{algpseudocode}
\usepackage{amsmath}
\usepackage{amsthm}
\usepackage{framed}
\usepackage{xspace}
\usepackage{warpcol}
\usepackage{color}

\usepackage{graphicx}
\usepackage{epsfig}
\usepackage{subfigure}
\usepackage{dcolumn}
\usepackage{float}
\usepackage[normalem]{ulem}
\useunder{\uline}{\ul}{}
\usepackage{multirow}
\usepackage{listings}
\usepackage{mathtools, amssymb}

\begin{document}

\begin{frontmatter}



\title{Uncertainty-Aware Deep Co-training for Semi-supervised Medical Image Segmentation}

\author[label1]{Xu Zheng}
\author{Chong Fu\fnref{label1,label2,label3} \corref{cor1}}
\ead{fuchong}
\cortext[cor1]{Corresponding author.}
\author[label1]{Haoyu Xie}
\author[label1]{Jialei Chen}
\author[label1]{Xingwei Wang}
\author[label4]{Chiu-Wing Sham}

\address[label1]{School of Computer Science and Engineering, Northeastern University, Shenyang 110819, China}
\address[label2]{Engineering Research Center of Security Technology of Complex Network System, Ministry of Education, China}
\address[label3]{Key Laboratory of Intelligent Computing in Medical Image, Ministry of Education, Northeastern University, Shenyang 110819, China}
\address[label4]{School of Computer Science, The University of Auckland, New Zealand}

\begin{abstract}
Semi-supervised learning has made significant strides in the medical domain since it alleviates the heavy burden of collecting abundant pixel-wise annotated data for semantic segmentation tasks. Existing semi-supervised approaches enhance the ability to extract features from unlabeled data with prior knowledge obtained from limited labeled data. However, due to the scarcity of labeled data, the features extracted by the models are limited in supervised learning, and the quality of predictions for unlabeled data also cannot be guaranteed. Both will impede consistency training. To this end, we proposed a novel uncertainty-aware scheme to make models learn regions purposefully. Specifically, we employ Monte Carlo Sampling as an estimation method to attain an uncertainty map, which can serve as a weight for losses to force the models to focus on the valuable region according to the characteristics of supervised learning and unsupervised learning. Simultaneously, in the backward process, we joint unsupervised and supervised losses to accelerate the convergence of the network via enhancing the gradient flow between different tasks. Quantitatively, we conduct extensive experiments on three challenging medical datasets. Experimental results show desirable improvements to state-of-the-art counterparts.

\end{abstract}




\end{frontmatter}


 \begin{figure*}[h]
	\centering
	\includegraphics[scale=0.2]{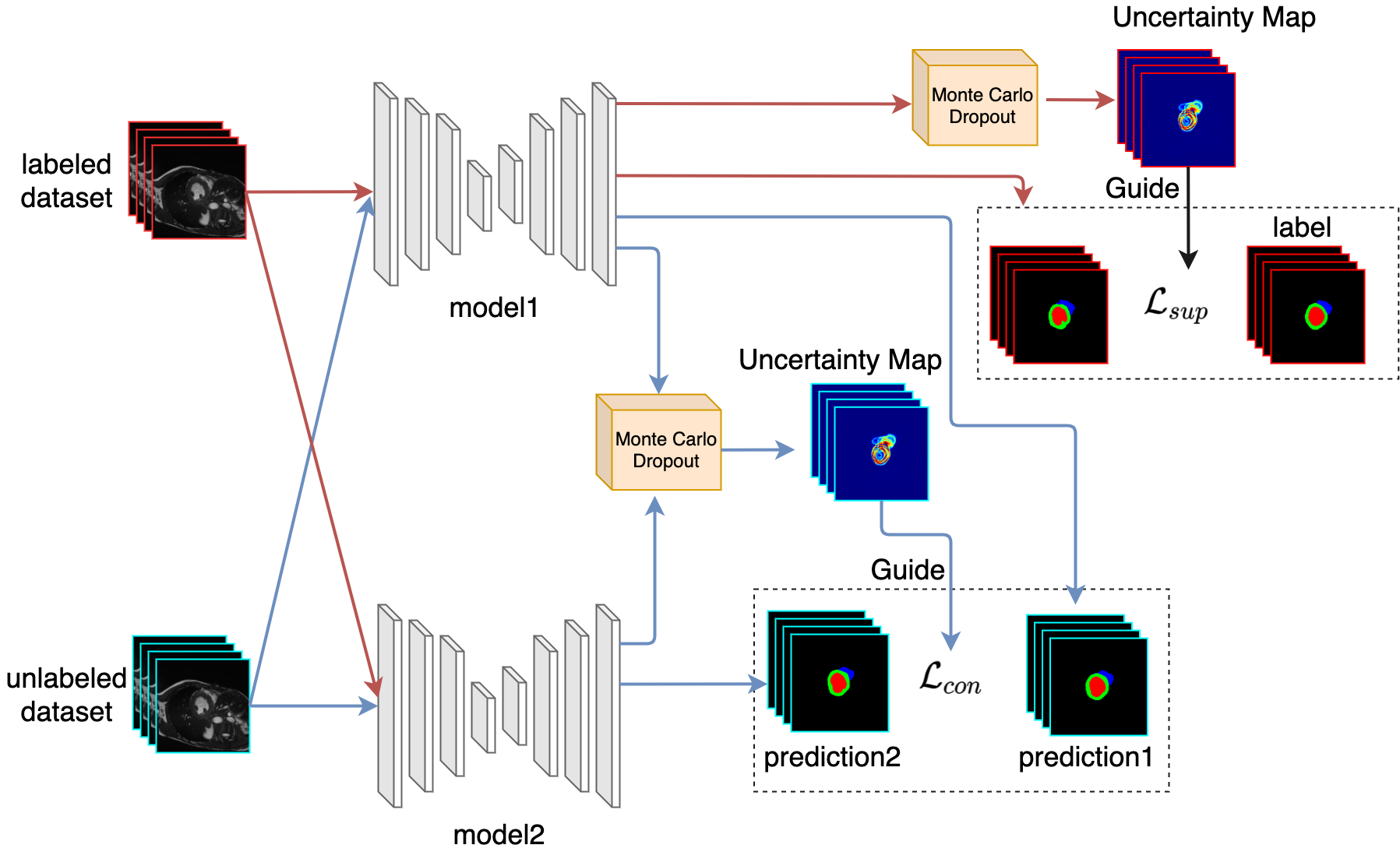}
	\caption{ Overview of the uncertainty-aware co-training algorithm. On the basis of training two deep CNN models simultaneously with two different sets of the labeled datasets and the same unlabeled dataset, we bring in Monte Carlo Sampling to acquire uncertainty, which is acting on the loss function in both the supervised learning stage and the unsupervised learning stage. 
	}
	\label{fig_framework}
\end{figure*}
\section{Introduction}
\label{intro}
Semantic segmentation is a fundamental computer vision task that is particularly important for medical image analysis~\cite{Medical-Image-Semantic-Segmentation}. It serves as a preprocessing step for the treatment of different medical conditions. In recent years, convolutional neural network (CNN) based methods have achieved tremendous improvement~\cite{deconvolutionnetworkforsemanticsegmentation},~\cite{Artificial-Convolutional-Neural-Network}. With abundant pixel-wise labeled data, CNN can easily achieve outstanding performance relying on great nonlinear fitting capability~\cite{Deep-semantic-segmentation}. However, pixel-wise annotated images are expensive to obtain especially in the medical domain. It is time-consuming and extremely relies on experienced experts. Fortunately, semi-supervised methods leverage not only labeled data but also unlabeled data~\cite{A-survey-on-semi}, freeing the researchers from labeling work. Therefore, semi-supervised learning has attracted great attention~\cite{Not-so-supervised},~\cite{Semi-Supervised-Crowd-Counting},~\cite{semi-supervised-plant}.

The current dominant semi-supervised learning methods in deep learning are self-training~\cite{Semi4FCN}, consistency regulation~\cite{self-ensembling}, pseudo labeling\cite{Pseudo-label}, adversarial learning~\cite{AdversarialLearning} and so on.  All the above approaches are based on the assumption that semi-supervised methods rely on intrinsic properties of the dataset distribution rather than individual images. So, parameter optimization can process with a combination of annotated images and unannotated images rather than labeled images only. For example, self-training~\cite{self-ensembling} selects the most reliable prediction of the current model as the label (pseudo label), for the complement of the labeled dataset. Other approaches attempt to exchange internal information among an ensemble of models, aimed at increasing their consistency, such as co-training~\cite{Deep-co-training}. These approaches are all constructive to a certain degree. But how to leverage unannotated data more effectively is still one of the most concerning issues in these semi-supervised learning methods. 

Most of the existing semi-supervised methods leverage the prediction of unannotated data, but the quality of the prediction is not guaranteed. Optimizing model parameters via unreliable predictions in unsupervised learning is not convinced, even towards wrong results. Here, the major challenge is to assess the reliability of the prediction. To address this issue, we propose an uncertainty-aware semi-supervised semantic segmentation framework, as illustrated in Fig.~1. Uncertainty is an estimation of prediction confidence~\cite{What-Uncertainties-Do-We-Need}, which is leveraged as prior information to guide network learning by adjusting loss weight dynamically in two separate strategies in our framework. It forces the models to focus on the valuable region according to the characteristics of supervised learning and unsupervised learning. On the one hand, it guides networks to pay more attention to the pixels with high prediction reliability (low uncertainty) in unsupervised learning; on the other hand, it guides networks to concentrate more on ambiguous pixels (high uncertainty) in supervised learning. In addition, owing to its success in semi-supervised learning, deep co-training is utilized in our proposed framework. The general principle of co-training is to independently train models with the labeled data and enforce their predictions to agree on the labeling of the unlabeled data. We separate the labeled dataset into complementary subsets and use adversarial examples~\cite{Explaining-adversarial-examples} to enforce the diversity. Moreover, to promote the predictions consistency among different models, we add ensemble agreement loss to guide models training. We joint unsupervised and supervised losses to accelerate the convergence of the network via enhancing the gradient flow between different tasks.

In this paper, we explore the uncertainty information to enforce the performance of networks both in supervised learning and unsupervised learning. With the guidance of the estimated uncertainty in both supervised and unsupervised stages, our uncertainty-aware deep co-training method achieves the state-of-the-art performance in the semi-supervised tasks on three public medical datasets. Moreover, a comprehensive set of experiments are conducted which demonstrate the potential of our approach for segmenting different types of medical images. Our experiments analyze the impact of various elements of the method. 
\section{Related work}
\label{sec2}
\subsection{Semi-supervised segmentation}
Semi-supervised learning is an indispensable part of deep learning theory~\cite{Semi-supervised-learning-literature-survey}. Recently, many works have proposed using semi-supervised approaches in medical image segmentation to segment human organs~\cite{Semi-supervised-learning-for-MR-image}. An innovative semi-supervised segmentation approach is presented by Mohamed et al. for efficient segmentation of lung CT scans from Convid-19 patients. Yu et al.~\cite{Uncertainty-aware} proposed a teacher-student model to segment 3D left atrium via self-ensembling. Li et al.~\cite{self-ensembling} proposed a semi-supervised segmentation method for skin lesions by enforcing the consistency between the student and teacher model. Fang et al.~\cite{Dmnet} used a co-training framework to boost each sub-model for kidney tumor segmentation. And they applied adversarial training~\cite{Adversarial-learning} to their network to constrain the models to output invariant results over different perturbations on input data.

Similar to co-training, Peng et al.~\cite{Deep-co-training} presented a method based on an ensemble of deep segmentation networks. They trained different models with corresponding subsets of the annotated data and leveraged non-annotated images to exchange internal information among sub-models.  The difference between these semi-supervised methods is how they utilize unlabeled data and the way they relate to supervised algorithms~\cite{A-survey-on-semi}. Most of the approaches utilize predictions of unlabeled data for consistency training~\cite{Mutual-Consistency-Training}.  However, the model's prediction of unlabeled data used to exchange information is unstable. We need to add components to enable us to learn the specific confidence situation of each pixel of the prediction~\cite{Confidence-Aware}. So we introduce uncertainty estimation~\cite{The-impact-of-uncertainty} in our approach.

\subsection{Uncertainty Estimation}
Uncertainty estimation plays a pivotal role in reducing the impact of stochastic during both optimization and decision-making processes~\cite{A-review-of-uncertainty-quantification}. Knowing the confidence (uncertainty) with which we can trust the neural networks' predictions is essential for decision making~\cite{Uncertainty-estimation}. A mathematically grounded framework is offered by Bayesian probability theory to learn about the uncertainty of a data-driven model, called Bayesian Neural Network(BNN).  BNN attempted to learn a distribution over each of the network's weight parameters~\cite{What-Uncertainties-Do-We-Need}. However, Bayesian inference is intractable on computation in practice. 

Deep ensemble~\cite{Deep-Ensembles} is an alternative to BNN, which observes the variance of many trained models to estimate uncertainty. In addition, several works applied different uncertainty estimation methods to the task of semantic segmentation, including Stochastic Batch Normalization~\cite{Uncertainty-estimation-via-stochastic}, Multiplicative Normalizing Flows~\cite{Multiplicative-normalizing-flows} and so on. 
Upadhyay et al.~\cite{Uncertainty-Aware-GAN} proposed a GAN-based framework that estimates the per-voxel uncertainty in the predictions on magnetic resonance imaging (MRI). Alex~\cite{What-Uncertainties-Do-We-Need} studied models under the framework with per-pixel semantic segmentation and proposed new loss functions based on uncertainty, interpreted as learned attenuation. Wu et al.~\cite{Mutual-Consistency-Training} claimed that deep models require extra components to obtain performance gains. And Yu et al.~\cite{Uncertainty-aware} introduced uncertainty to ensemble models by using Monte-Carlo Sampling. It has been proved that the Monte Carlo Sampling has much superiority over conventional methods in the estimation of uncertainty, especially of complex measurement systems' output~\cite{Uncertainty-estimation}.
\section{Methods}
\subsection{Problem formulation}
Since manual annotation is often time-consuming and expensive, only a small fraction of images in datasets can have full pixel-wise labels. So our proposed method aims to learn more information from both labeled and abundant unlabeled images by using the uncertainty-aware deep co-training methodology.
We formalize the problem of medical image semantic segmentation as follows.

Given labeled dataset $D_l={\{(x_1^l, y_1^l), ... , (x_m^l, y_m^l)}\}$, which contains $m$ labeled examples, each example comprised of an input image $x_i^l$ with dimensions ${H\times W}$ and its corresponding pixel-level label $y_i^l \in \mathbb{R}^{H\times W \times C}$, where $C$ is the number of classes and $H\times W$ is spatial dimension . In semi-supervised setting, we also have unlabeled dataset $D_u=\{x_1^u, ..., x_n^u\} $ including $n$ unlabeled images, with $n \gg m$. The purpose of semi-supervised segmentation is to train a segmentation model $f$ of parameter $\theta$ with $D = D_l \cup D_u$, which map each pixel of an input image to its correct label.
\begin{figure}[]
	\centering
	\includegraphics[scale=0.25]{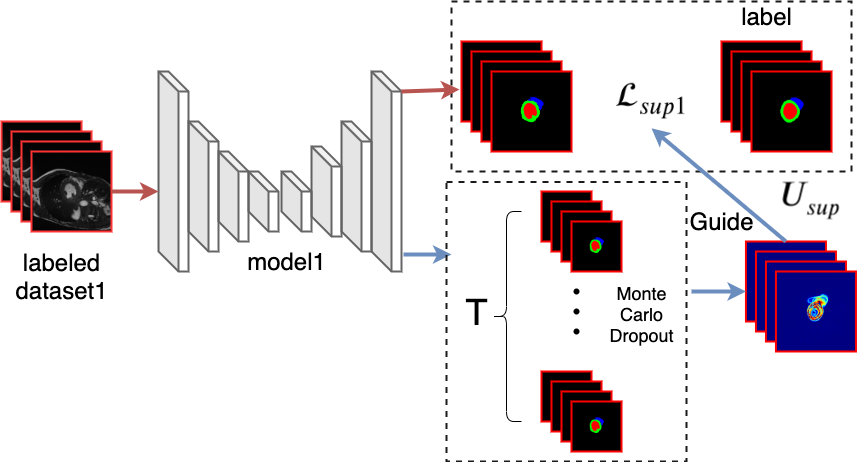}
	\caption{Illustration of supervised learning. Each model outputs predictions of the image from corresponding labeled subsets, computing supervised loss $L_{sup}$ with labels, illustrated as red lines. Simultaneously, the uncertainty is estimated by Monte Carlo Sampling. Each model approximates its own uncertainty by performing $T$ stochastic forward passes and multiplies the uncertainty $U_{sup}$ as a weight coefficient to the supervised loss $L_{sup}$.}
	\label{}
\end{figure}
\subsection{Proposed approach}
We train two models $f(\theta_1)$, $f(\theta_2)$ jointly with labeled dataset $D_l$ and unlabeled dataset $D_u$ in a collaborative manner. Motivated by performance improvement from the uncertainty estimation in the Bayesian networks, we employ Monte Carlo Sampling\cite{What-Uncertainties-Do-We-Need} to estimate uncertainty. Specifically, we perform $T$ stochastic forward passes on two models under random dropout. According to this, we obtain a set of predictions and choose the predictive entropy metric to approximate the uncertainty. The predictive entropy can be summarized as: 
\begin{equation}
	\begin{aligned}
		\mu_c= \frac{1}{T}\sum_{t}^{}P_{t}^{c} \qquad  and\qquad u=-\sum_{c}^{}\mu_{c}\mathit{log}\mu_{c},
	\end{aligned}
\end{equation}
where $P_{t}^{c}$ is the probability of the c-th class in the t-th time prediction. And the uncertainty is estimated in pixel level, the uncertainty of the whole volume $U$ is $ \left \{u  \right \}\in \mathbf{R}^{H\times W\times D}$.

Following co-training methods for semi-supervised semantic segmentation, we employ a loss function composed of a weighted sum of three separate terms to train ensemble models. \\
\begin{equation}
	\mathcal{L}(\theta;D)  = \mathcal{L}_{sup}(\theta;D_l) + \lambda _{cot}\mathcal{L}_{agr}(\theta ;D_u) + \lambda _{div}\mathcal{L}_{div}(\theta ;D).
\end{equation}
Uncertainty is used to impact the weight of the first two parts of the loss function. The details are explained in the following subsections.

\subsubsection{Supervised learning}
As in standard multi-view learning~\cite{Deep-multi-view-learning} approach, we separate the labeled dataset $D_l$ to complementary subsets $D_{l_1}$ and $D_{l_2}$. Training two models individually with corresponding labeled subsets makes each model have a certain fit to the distribution of the data. And also, we can ensure the diversity of the two models.

We employ E-Net\cite{enet} as our network backbone. To adapt the E-Net as a Bayesian network to estimate the uncertainty, one dropout layer is added between the encoder and decoder. In the network training and uncertainty estimation, we turn on the dropout layer. Since we don't need to estimate uncertainty in the testing phase, we turn off the dropout layer.

Through the forward propagation of each model $T$ times, we can get $T$ different results as shown in Fig.~2. Therefore, we obtain a set of Softmax probability vectors for each pixel in the input. By using predictive entropy as the metric, we can approximate uncertainty maps. We express the uncertainty map as the model confidence of each pixel prediction. If a pixel has a large fluctuation during the prediction process of $T$ forward propagation, the uncertainty of it will be large.

With the guidance of the estimated uncertainty $U_{sup_1}$ and $U_{sup_2}$, we find out the unreliable prediction in images and make models learn more from these pixels from corresponding labels. We design the uncertainty-aware supervised loss $L_{sup}$ as the pixel-level cross-entropy loss of the two models:
\begin{equation}
	\centering
	\begin{aligned}
		\mathcal{L}_{sup}(\theta ,D_l)=U_{sup_1}\sum_{D_{l_1}}^{}\mathit{CE}(f(x_i;\theta_{1}),y_i)\\
		+U_{sup_2}\sum_{D_{l_2}}^{}\mathit{CE}(f(x_i;\theta_{2}),y_i).	
	\end{aligned}
\end{equation}

\subsubsection{Unsupervised learning}
Semi-supervised learning means not only exploiting labeled images, unlabeled images $D_u$ are more important for the learning process. The strategy of unsupervised learning wants the segmentation networks to output similar predictions for the same unlabeled image with perturbations. The deep co-training method minimizes the distance between the class distributions predicted by different models. However, the prediction results obtained through one forward pass are contingent, which will limit models' ability to fit the unlabeled data distribution.

So, we also introduce the concept of uncertainty in unsupervised learning. Same as the fully supervised phase, we perform $T$ stochastic forward passes on each model under random dropout as shown in Fig.~3. After leveraging predictive entropy as a metric, we get two uncertainty maps approximated by models. Different from the supervised stage, we combine the results of the two models here. We calculate the average of the two uncertainties and take it as a weight coefficient and multiply it on the loss function. Since the calculated uncertainty fluctuates greatly, we normalize it in this way: 
\begin{equation}
	U_{un} = -\beta \times [0.5(U_{un1} + U_{un2}) + C],
\end{equation}
where $\beta$ and $C$ are constants used for normalization. In order t to make the uncertainty change within the controllable range, we set $\beta = 0.7$, $C = 2$.

In supervised learning, the models should pay more attention to the pixels with higher uncertainty. But here in unsupervised learning, models should learn more from lower uncertainty pixels as there is no label in this stage. The estimated uncertainty guides models concentrating more on reliable (low uncertainty) predictions. We add uncertainty $U_{un_1}$, $U_{un_2}$ estimated by two models together, and normalize it as $U_{un}$. The uncertainty-aware ensemble agreement loss we leverage here is Jensen-Shannon divergence:
\begin{equation}
	\centering
	\begin{aligned}
		\mathcal{L}_{agr}=U_{un}\mathbb{E}_{D_u}[D_{JS}(f(x_{i}^{u};{\theta_1})||f(x_{i}^{u};{\theta_2}))].	
	\end{aligned}
\end{equation}

\begin{figure}[]
	\centering
	\includegraphics[scale=0.14]{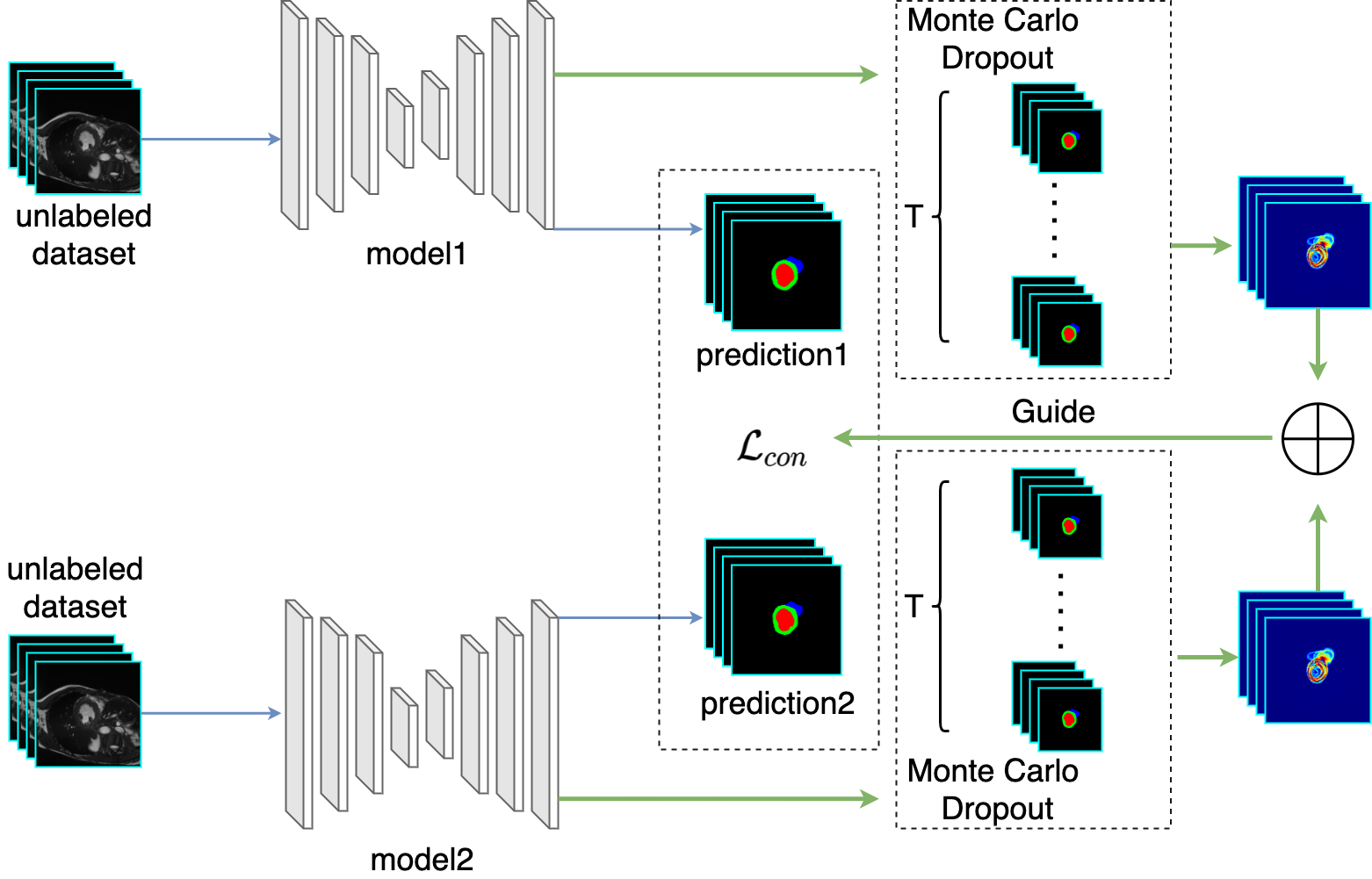}
	\caption{Illustration of unsupervised learning. Two networks output predictions based on the knowledge from each view. We leverage the ensemble agreement loss $\mathcal{L}_{con}$ to encourage the predictions of models to consistency with the guidance of the estimated uncertainty. The final uncertainty is the sum of the uncertainties generated by two models using Monte Carlo Dropout. Each model performs $T$ stochastic forward passes to get an uncertainty map.
	}
	\label{fig_framework}
\end{figure}
\begin{figure*}
	\centering
	\begin{minipage}{0.75\linewidth}
		\includegraphics[width=0.5\textwidth]{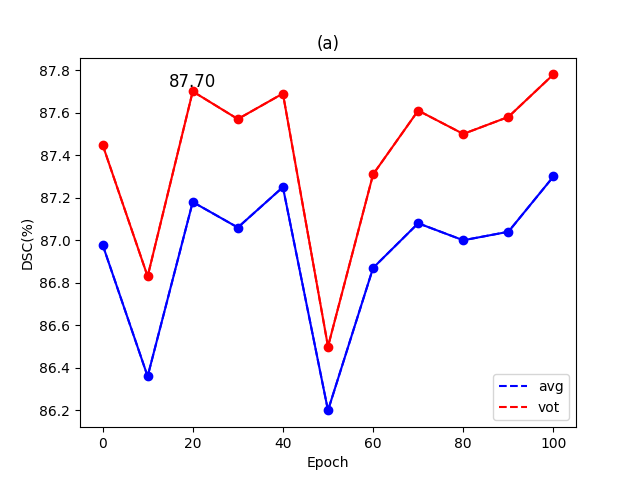}
		\includegraphics[width=0.5\textwidth]{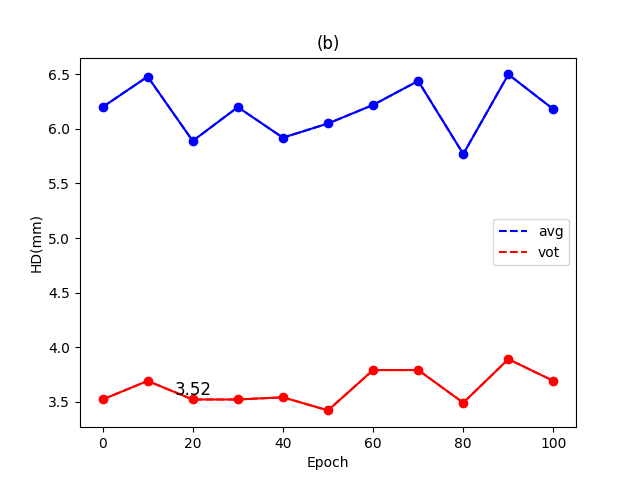}
	\end{minipage}
	\begin{minipage}{0.75\linewidth}
		\includegraphics[width=0.5\textwidth]{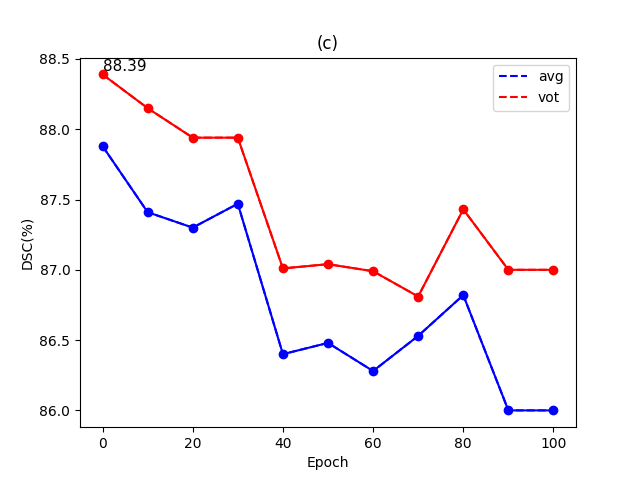}
		\includegraphics[width=0.5\textwidth]{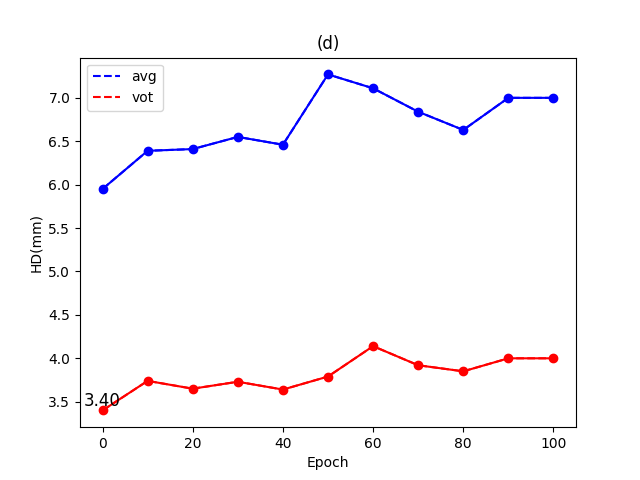}
	\end{minipage}
	\centering
	\caption{Illustration of the DSC(\%) and HD(mm) results of introducing uncertainty in different epochs in the supervised and unsupervised stages. Pictures (a), (b) represent leveraging uncertainty only in the unsupervised stage, and (c), (d) represent using uncertainty only in the supervised stage. The red broken lines in each picture represent voting results by two models, and the blue broken lines represent average results. Considering the performance both in DSC and HD, we finally choose to add uncertainty at 0 epoch of supervised stage, and at 20 epoch of unsupervised stage.}
	\vspace{-0.2cm}
	\label{fig1}
\end{figure*}
\subsubsection{When to introduce uncertainty? }
Introducing uncertainty into models can improve the performance of the network, but introducing it at unappropriate time will crush the entire framework. 

In the supervised stage, theoretically, the models don't need prior knowledge to avoid turning in the wrong direction. So the sooner uncertainty is introduced to guide network learning, the better performance we can get. We verify this by doing experiments of taking uncertainty in concern at different epoch. The results are shown in Fig.~4. Considering the performance in both DSC and HD, we leverage uncertainty from the beginning in the supervised stage.

In the unsupervised stage, the models' prediction error for unlabeled images is large at the beginning of experiments. This will lead to a large fluctuation range of uncertainty, which is not conducive to allowing the model to focus on where it should learn more. So, we tested adding uncertainty to the unsupervised stage at different epoch, and the final results are shown in Fig.~4. And finally, we learned that introducing uncertainty in the unsupervised stage requires that models have a certain understanding of the target distribution.

Synthesize all factors, we introduce uncertainty to the supervised stage at epoch 0, and epoch 20 for the unsupervised stage in all experiments of our approach. 

\subsubsection{Adversarial learning}
Having diversity between models in the ensemble is indispensable. In co-training, diversity is essential for avoiding the collapse of decision boundary and the models can learn from each other during training. The deep co-training method uses the approach proposed by Qiao et al. for image classification and augments the dataset with adversarial examples. The adversarial examples are generated from both labeled and unlabeled data.

One model uses adversarial examples teaching other models in the ensemble. In the original deep co-training method, the diversity loss is:
\begin{equation}
	\centering
	\begin{aligned}[]
		\mathcal{L}_{div} (\theta;D) = \mathbb{E}_{x\in D}[\mathcal{H}(f^1(x;\theta^1),f^2(g^1(x);\theta^2)) \\ +\mathcal{H}(f^2(x;\theta^2),f^1(g^2(x);\theta^1))],
	\end{aligned}
\end{equation}
where $\mathcal{H}(.)$ refers to cross entropy and $x$ is an input image. The $g^i(x)$ is an adversarial example target on model $f^i(. ;\theta^i)$. So $g^1(x)$ is an adversarial example for model 1, and we can easily get that $f^1(x;\theta^1)$ $\not =$ $f^1(g^1(x);\theta^1)$. In the process of minimizing the first term of \textcolor{blue}{Eq.~6.}, it will make $f^1(x;\theta^1)$  approximately equal to $f^2(g^1(x);\theta^2)$. Combining the above relations, we can obtain $f^1(g^1(x);\theta^1)$  $\neq$ $f^2(g^1(x);\theta^2)$. Applying the same idea to model 2, we can conclude that the models have divergence in predicting adversarial examples.

Adversarial examples can be obtained by adding small perturbations to input images. As described in ~\cite{Deep-co-training}, we follow use different schemes to generate these examples based on the source of the image $x$. We apply the Virtual Adversarial Training (VAT) method when $x$ is an unlabeled image from $D_{u}$; and we use the Fast Gradient Sign Method (FGSM) when $x$ in labeled datasets. 

\begin{table*}[h]
	\centering
	\caption{Comparison of semi-supervised method results on segmentation from ACDC dataset with $l_{\alpha}$=0.2. All the experiments are reimplemented by us.}\label{tbl2}
	\begin{tabular}{llllll}
		\hline
		\multicolumn{2}{l}{\multirow{2}{*}{Method}} & \multicolumn{4}{c}{DSC($\%$)} \\ \cline{3-6} 
		\multicolumn{2}{l}{}                        & RV   & Myo   & LV   & Mean  \\ \hline
		Full                  &                     &$89.33(0.23)$      &$88.07(0.53)$       &$93.34(0.48)$      &$90.25(0.48)$    \\ 
		MT                    &                     &$76.70(7.07)$      &$74.26(4.78)$       &$84.73(1.24)$      &$78.56(2.86)$       \\ 
		UAMT                  &                     &$77.69(1.04)$      &$78.88(1.21)$       &$88.17(0.57)$      &$81.58(0.32)$       \\ 
		Part                  & avg                 &$73.89(1.90)$      &$75.64(2.25)$       &$86.95(0.85)$     &$78.83(1.66)$       \\ 
		& vot                 &$75.16(2.00)$      &$77.61(2.39)$       &$88.38(0.89)$      &$80.39(1.72)$       \\ 
		DCT                   & avg                 &$81.89(1.95)$      &$83.32(0.78)$      &$91.13(0.45)$      &$85.45(0.68)$       \\ 
		& vot                 &$82.45(2.17)$      &$84.35(0.58)$       &$91.68(0.62)$      &$86.16(0.73)$       \\ 
		Ours                  & avg                 &$85.91(0.12)$      &$85.96(0.27)$       &$92.34(0.26)$      &$88.07(0.15)$       \\ 
		& vot                 &$\textbf{86.59(0.17)}$      &$\textbf{86.65(0.35)}$       &$\textbf{92.74(0.33)}$      &$\textbf{88.45(0.12)}$       \\ \hline
		\multicolumn{2}{l}{\multirow{2}{*}{Method}} & \multicolumn{4}{c}{HD($mm$)} \\ \cline{3-6} 
		\multicolumn{2}{l}{}                        & RV   & Myo   & LV   & Mean  \\ \hline
		Full                  &                     &$3.77(0.15)$      &$3.09(0.06)$       &$2.40(0.10)$      &$3.09(0.03)$       \\ 
		MT                    &                     &$10.03(1.07)$      &$10.47(3.31)$      &$10.34(4.09)$      &$10.28(2.25)$       \\ 
		UAMT                  &                     &$8.05(0.93)$      &$7.18(3.31)$       &$6.30(2.18)$     &$7.18(1.12)$      \\ 
		Part                  & avg                 &$14.09(0.52)$      &$12.14(0.80)$       &$8.87(0.74)$      &$11.70(0.42)$       \\ 
		& vot                 &$7.49(0.23)$       &$6.25(0.43)$      &$4.57(0.19)$      &$6.11(0.24)$      \\ 
		DCT                   & avg                 &$8.16(0.93)$      &$7.42(1.65)$       &$5.38(0.12)$      &$7.23(0.48)$       \\ 
		& vot                 &\textbf{4.84(0.57)}      &$4.06(0.19)$       &$3.01(0.08)$      &$3.97(0.22)$       \\ 
		Ours                  & avg                 &$6.74(0.33)$      &$6.34(0.48)$       &$4.60(0.30)$      &$5.89(0.25)$      \\ 
		& vot                 &$4.93(1.30)$      &$\textbf{3.65(0.18)}$       &$\textbf{2.68(0.08)}$      &$\textbf{3.48(0.05)}$       \\ \hline
	\end{tabular}
\end{table*}
\begin{table*}[]
	\centering
	\caption{Comparison of semantic segmentation results with different label ratios in ACDC dataset.}\label{tbl1}
	\begin{tabular}{cccccccc}
		\hline
		\multicolumn{2}{c}{\multirow{2}{*}{Method}} & \multicolumn{6}{c}{DSC(\%)}                                                                         \\ \cline{3-8} 
		\multicolumn{2}{c}{}                        & $5$              & $10$            & $20$             &$30$            & $40$             & $50$             \\ \hline
		\multicolumn{2}{c}{Part}                    & $69.39$          & $76.34$          & $79.12$          &$88.06$          & $89.11$          & $89.67$          \\ 
		\multirow{2}{*}{IND}          & avg         & $73.26$          & $73.59$          & $81.16$          & $86.16$          & $86.64$          & $86.67$          \\ 
		& vot         & $73.65$          & $75.66$          & $82.82$          & $87.30$          & $88.04$          & $88.68$          \\ 
		\multirow{2}{*}{DCT}          & avg         & $79.45$          & $85.67$          & $85.24$          & $87.28$          & $88.37$         & $88.19$          \\ 
		& vot         & $80.43$          & $86.40$          & $86.31$          & $88.02$          & $89.01$          & $89.10$          \\
		\multirow{2}{*}{Ours}         & avg         & $82.52$          & $87.58$          & $87.99$          & $88.93$          & $89.37$          & $90.40$          \\ 
		& vot         & $\textbf{83.16}$ & $\textbf{88.01}$ & $\textbf{88.57}$ & $\textbf{89.41}$ & $\textbf{89.86}$ & $\textbf{90.89}$ \\ \hline
		\multicolumn{2}{c}{\multirow{2}{*}{Method}} & \multicolumn{6}{c}{HD(mm)}                                                                          \\ \cline{3-8} 
		\multicolumn{2}{c}{}                        & 5              & 10             & 20             & 30             & 40             & 50             \\ \hline
		\multicolumn{2}{c}{Part}                    & $9.29$           & $6.53$           & $12.87$          & $4.01$           & $3.44$           & $3.29$           \\ 
		\multirow{2}{*}{IND}          & avg         & $12.14$          & $16.13$          & $12.28$          & $7.97$           & $7.68$           & $6.94$           \\  
		& vot         & $5.59$           & $8.25$           & $5.82$           & $4.43$           & $4.27$           & $3.83$           \\
		\multirow{2}{*}{DCT}          & avg         & $11.53$          & $6.86$           & $7.89$           & $6.06$           & $5.73$           & $5.93$           \\ 
		& vot         & $6.04$           & $4.11$           &$4.27$           & $3.58$           & $3.38$           & $3.62$           \\
		\multirow{2}{*}{Ours}         & avg         & $7.59$           & $5.53$           & $6.01$           & $5.33$           & $5.07$           & $4.89$           \\
		& vot         & $\textbf{4.70}$  &$ \textbf{3.42}$  & $\textbf{3.54}$  & $\textbf{3.09}$  & $\textbf{3.15}$  & $\textbf{2.78}$  \\ \hline
	\end{tabular}
\end{table*}
\section{Experiments and results}
\subsection{Dataset and metrics}
We evaluated our method on three medical image segmentation benchmark publicly available datasets: Automated Cardiac Challenge (ACDC)\cite{Automatic-MRI-Cardiac}, Spinal Cord Gray Matter Challenge (SCGM)\cite{Spinal-cord-grey-matter}, and Spleen sub-task dataset of the Medical Segmentation Decathlon Challenge\cite{A-large-annotated-medical-image-dataset}.
\paragraph{\textbf{ACDC dataset}}: 
The University Hospital of Dijon created the ACDC dataset from real clinical exams. The ACDC dataset covers several well-defined pathologies with enough cases. It consists of 200 short-axis cine-MRI scans from 100 patients. This dataset is divided into 5 evenly distributed medical groups: patients without cardiac disease, myocardial infarction, hypertrophic cardiomyopathy, abnormal right ventricles, and dilated cardiomyopathy. There are four regions of interest in segmentation masks: right ventricle endocardium (RV), left ventricle endocardium (LV), left ventricle myocardium (Myo), and background. We leveraged 75 subjects (150 scans) for training and 25 subjects (50 scans) for testing. All short-axis slices within 3D-MRI scans were resized to 256 $\times$ 256 as 2D images.
\paragraph{\textbf{SCGM dataset}}: 
The  Spinal Cord Gray Matter Challenge was organized to test the different capabilities of various methods. 
The dataset in this challenge is a public-available collection of multi-vendors, multi-center MRI. All data was acquired at 4 different sites: University College London, Polytechnique Montreal, University of Zurich, and Vanderbilt University. It contains 80 healthy subjects (age range of 28.3 to 44.3 years) with 20 subjects from each center.  The training set contains 40 labeled scans, each annotated slice-wise by 4 independent experts. The ground truth mask is obtained by majority voting. In our work, we leverage 30 images from the first center as a labeled dataset, and 465 images from all centers as an unlabeled dataset. The test set contains 264 labeled images from center 3 and center 4. The slices are center-cropped to 200 $\times$ 200 pixels.
\paragraph{\textbf{Spleen dataset}}: Spleen dataset is one of the ten sub-tasks of the Medical Segmentation Decathlon Challenge\cite{A-large-annotated-medical-image-dataset}. It consists of patients undergoing chemotherapy treatment for liver metastases as a publicly available dataset. The dataset includes 61 portal venous phases CT scans (only 41 were given with ground truth). The ground truth segmentation was generated by a semi-automatic segmentation software and identified by an expert abdominal radiologist. Each slice obtained from the CT scans is resized to 256 $\times$ 256 pixels. We split the dataset into labeled (4 patients), unlabeled (32 patients), and validation image subsets( 5 patients).
\paragraph{\textbf{Dice similiarity coefficient(DSC)}}: DSC measures the overlap between the predictions of the model $S$ and the ground truth $G$:
\begin{equation}
	DSC(S,G) = \frac{2|S\cap G|}{|S|+|G|}.
\end{equation}
\paragraph{\textbf{Hausdorff distance(HD)}}: HD is a boundary distance metric which measures the largest distance (in mm) between a point in prediction $S$ and the closest point in the ground truth $G$:
\begin{equation}
	HD(S,G) = max \begin{Bmatrix}d(S,G),d(G,S)\end{Bmatrix}.
\end{equation}
\begin{figure*}[h]
	\includegraphics[width=1\textwidth]{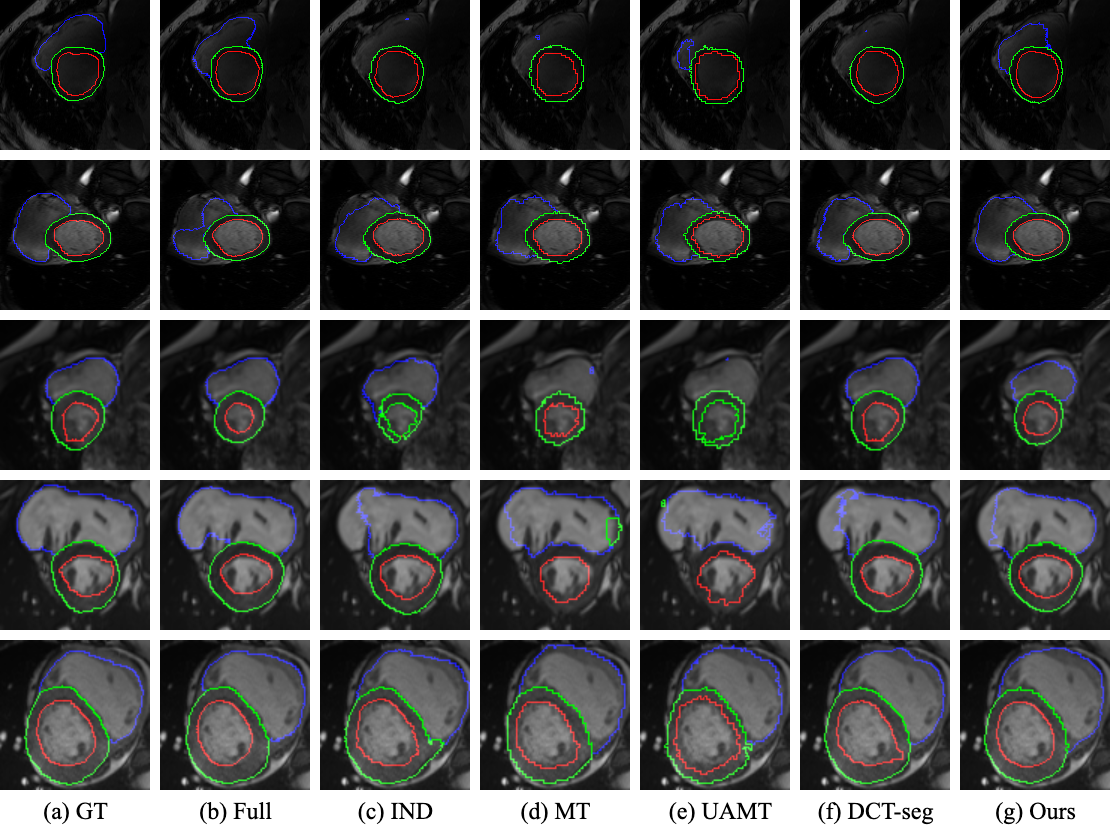}\\
	\centering
	\caption{Examples of segmentation results for ACDC dataset with 20$\%$ label images. From left to right: Ground truth (GT), Fully supervised baseline (Full), Independent, Mean Teacher (MT), Uncertainty-aware Mean Teacher (UAMT), Deep co-training (DCT), and our Uncertainty-aware Deep co-training method (Ours). }
	\vspace{-0.2cm}
	\label{fig1}
\end{figure*}
\begin{figure*}[h]
	\centering
	\includegraphics[width=1\textwidth]{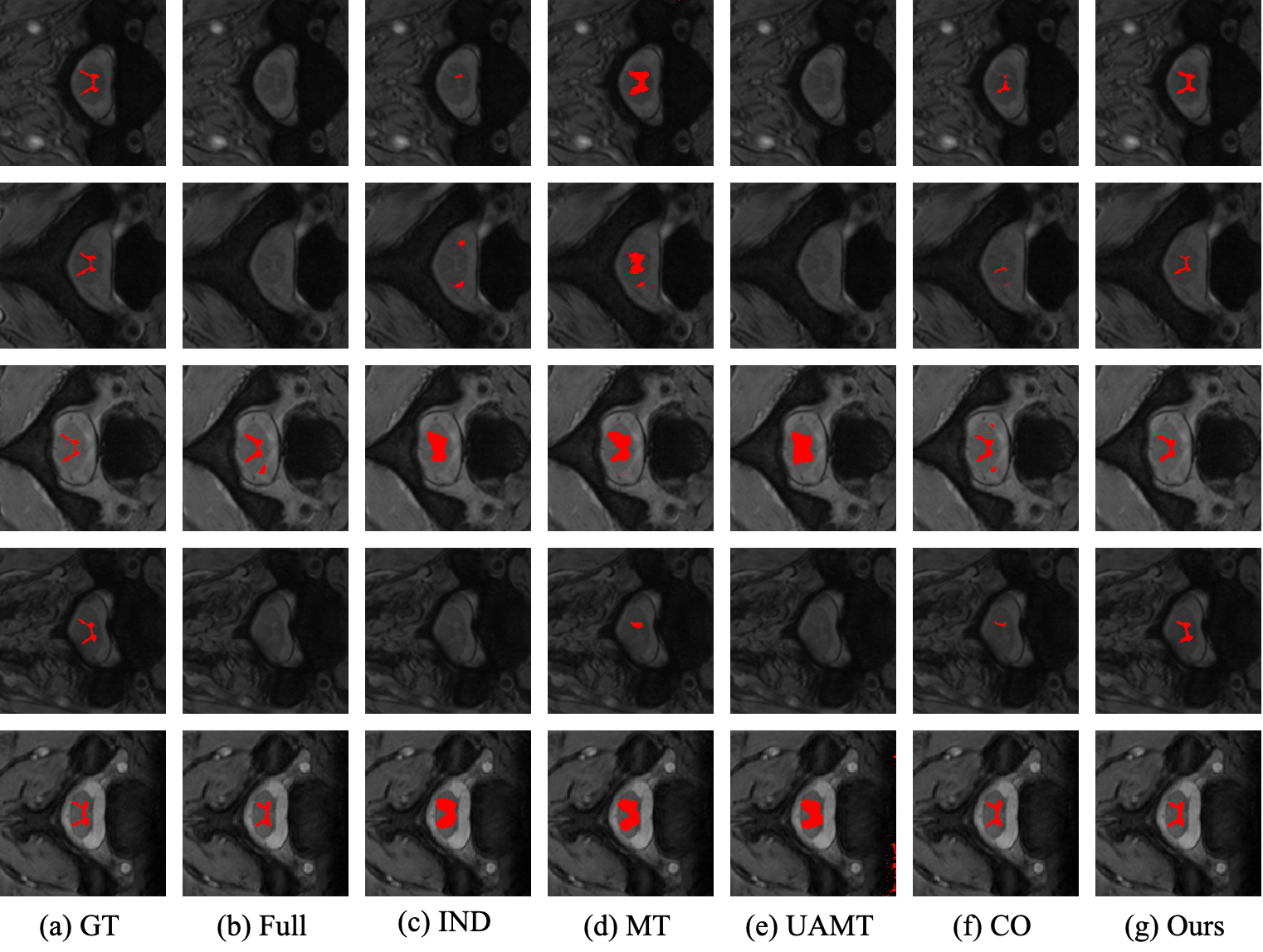}\\
	\centering
	\caption{Examples of segmentation results for SCGM dataset using center 1 as training data. From left to right: Ground truth (GT), Fully supervised baseline (Full), Independent, Mean Teacher (MT), Uncertainty-aware Mean Teacher (UAMT), Deep co-training (DCT), and our Uncertainty-aware Deep co-training method (Ours). }
	\vspace{-0.2cm}
	\label{fig1}
\end{figure*}
\begin{figure*}[h]
	\centering
	\includegraphics[width=1\textwidth]{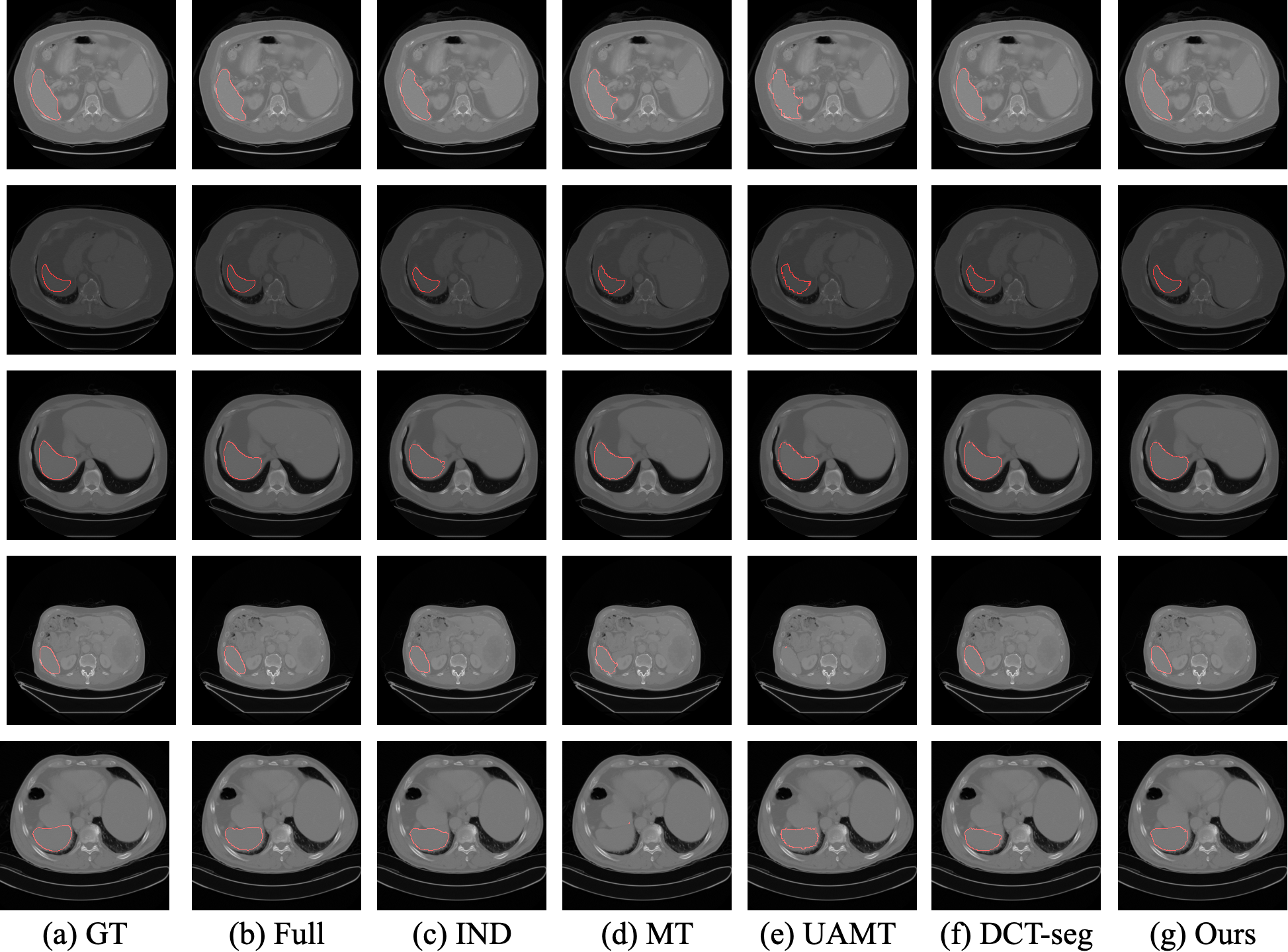}\\
	\centering
	\caption{Examples of segmentation results for Spleen dataset with 20\% label images. From left to right: Ground truth (GT), Fully supervised baseline (Full), Independent, Mean Teacher (MT), Uncertainty-aware Mean Teacher (UAMT), Deep co-training (DCT), and our Uncertainty-aware Deep co-training method (Ours).}
	\vspace{-0.2cm}
	\label{fig1}
\end{figure*}
\begin{figure*}[]
	\centering
	\includegraphics[width=0.85\textwidth]{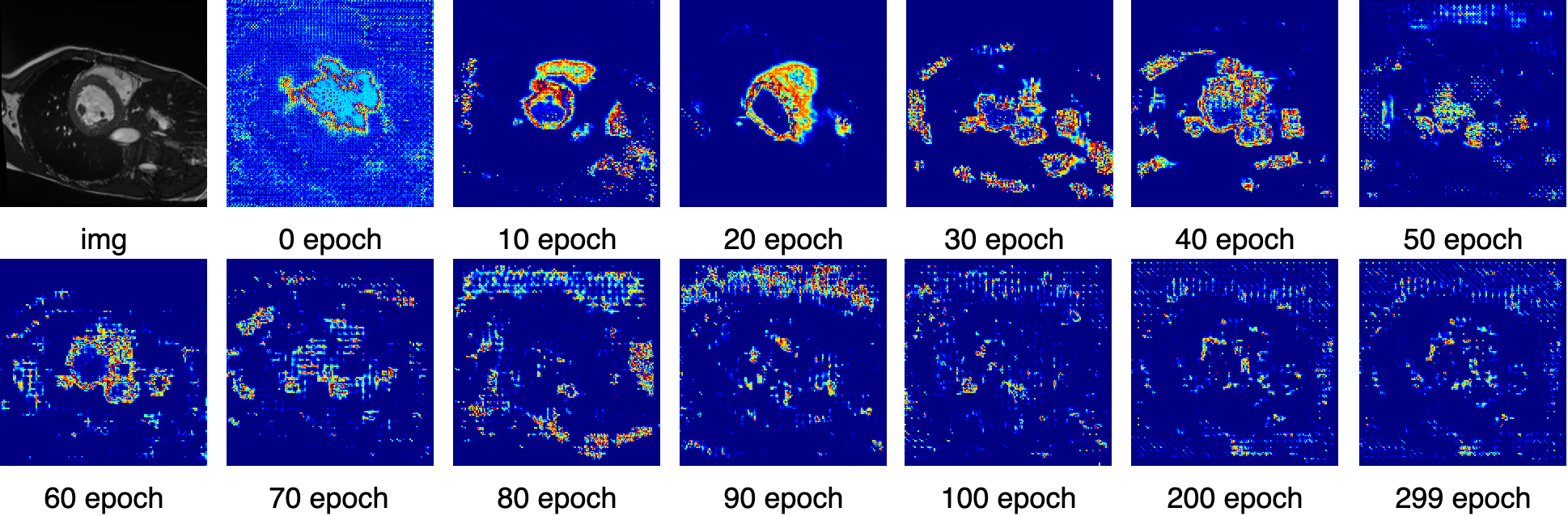} 
	\caption{Uncertainty heatmap under different epochs in supervised learning.}
\end{figure*}
\begin{figure*}[]
	\centering
	\includegraphics[width=0.85\textwidth]{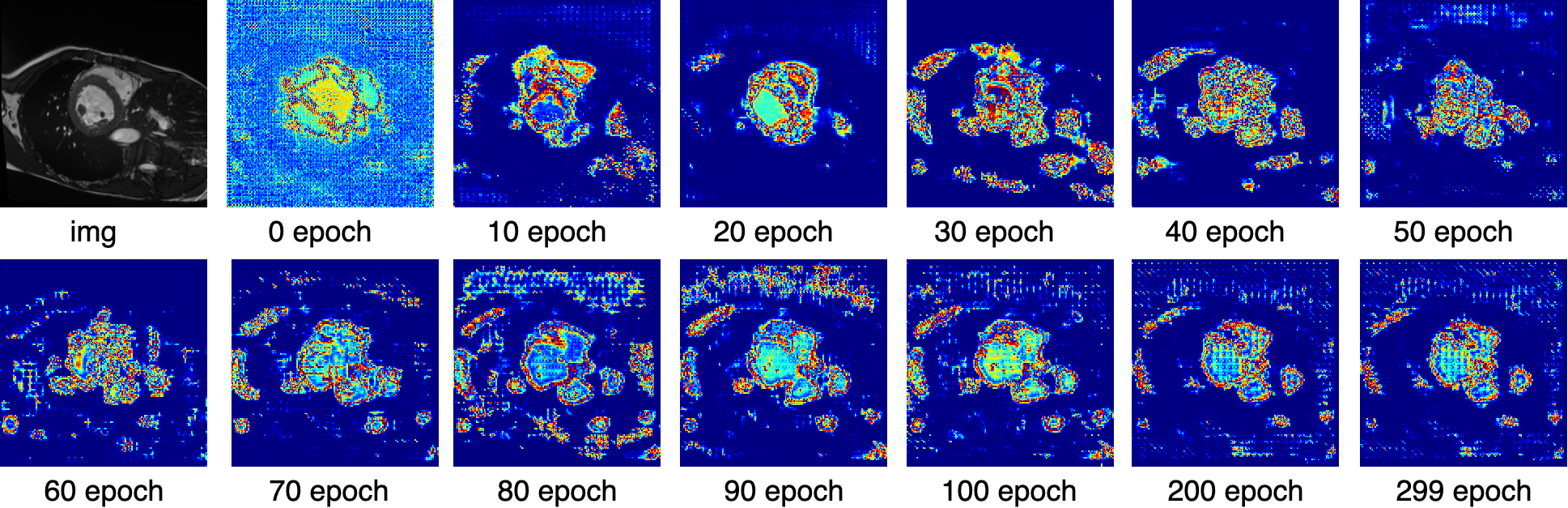} 
	\caption{Uncertainty heatmap under different epochs in unsupervised learning.}
\end{figure*}
\begin{table}[]
	\caption{Comparison of semi-supervised method results on segmentation from SCGM dataset.}\label{tbl3}
	\begin{tabular}{cccc}
		\hline
		\multicolumn{2}{c}{Method}         & DSC($\%$)     & HD($mm$)       \\ \hline
		\multicolumn{2}{c}{Full}           & $61.37(3.05)$ & $32.27(13.52)$ \\
		\multicolumn{2}{c}{Mean Teacher}   & $49.13(1.07)$ & $16.72(2.25)$  \\ 
		\multicolumn{2}{c}{UAMT}           & $39.82(2.28)$ & $39.35(6.04)$  \\ 
		\multirow{2}{*}{Independent} & avg & $48.09(0.65)$ & $33.42(2.01)$  \\
		& vot & $47.49(0.12)$ & $37.23(4.10)$  \\
		\multirow{2}{*}{DCT}     & avg & $66.02(1.92)$ & $18.56(2.21)$  \\ 
		& vot & $68.35(1.97)$ & $15.36(3.85)$  \\
		\multirow{2}{*}{Ours}        & avg & $73.19(0.25)$ & $10.05(1.05)$  \\  
		& vot & $\textbf{74.31(0.30)}$ & $\textbf{9.01(0.73)}$   \\ \hline
	\end{tabular}
\end{table}
\begin{table}[]
	\caption{Comparison of semi-supervised method results on segmentation from Spleen dataset.}\label{tbl4}
	\begin{tabular}{cccc}
		\hline
		\multicolumn{2}{c}{Method}         & DSC($\%$)     & HD($mm$)       \\ \hline
		\multicolumn{2}{c}{Full}           & $94.56(0.19)$ & $6.84(0.52)$ \\
		\multicolumn{2}{c}{Mean Teacher}   & $82.25(1.45)$ & $9.38(1.58)$  \\ 
		\multicolumn{2}{c}{UAMT}   & $84.07(1.02)$ & $15.48(4.46)$  \\ 
		\multirow{2}{*}{Independent} & avg & $91.99(0.35)$ & $12.44(1.36)$  \\
		& vot & $92.85(0.29)$ & $10.75(0.32)$  \\
		\multirow{2}{*}{DCT}     & avg & $93.08(0.60)$ & $8.88(1.73)$  \\ 
		& vot & $94.09(0.58)$ & $6.86(0.61)$  \\
		\multirow{2}{*}{Ours}        & avg & $94.55(0.19)$ & $7.24(0.14)$  \\  
		& vot & $\textbf{95.45(0.24)}$ & $\textbf{5.57(0.21)}$   \\ \hline
	\end{tabular}
\end{table}
\begin{table}[h]
	\caption{Comparison of different label ratios in Spleen dataset.}\label{tbl6}
	\begin{tabular}{cccc}
		\hline
		\multicolumn{2}{c}{\multirow{2}{*}{Method}} & \multicolumn{2}{c}{DSC}   \\ \cline{3-4} 
		\multicolumn{2}{c}{}                        & $5\%$         & $10\%$        \\ \hline
		\multirow{2}{*}{Independent}      & avg     & $84.51(0.97)$           & $88.18(0.42)$ \\ 
		& vot     & $86.94(0.61)$             & $89.79(0.73)$ \\ 
		\multirow{2}{*}{DCT}          & avg     & $86.77(2.63)$ & $88.56(0.58)$ \\ 
		& vot     & $87.91(2.63)$ & $90.66(0.27)$ \\ 
		\multirow{2}{*}{Ours}             & avg     & $87.76(1.59)$ & $90.06(0.51)$ \\  
		& vot     & $\textbf{89.02(1.05)}$ & $\textbf{91.94(0.77)}$ \\ \hline
	\end{tabular}
\end{table}

\begin{table*}[h]
	\caption{Abalation analysis of our method. }\label{tbl6}
	\begin{tabular}{clllllllll}
		\hline
		\multicolumn{2}{c}{\multirow{2}{*}{Method}}  & \multicolumn{4}{c}{DSC(\%)}   & \multicolumn{4}{c}{HD(mm)} \\ \cline{3-10} 
		\multicolumn{2}{c}{}                         & RV    & Myo   & LV    & Mean  & RV    & Myo  & LV   & Mean \\ \hline
		\multirow{2}{*}{DCT-seg}               & avg & 81.76 & 82.21 & 91.75 & 85.24 & 9.31  & 9.16 & 5.21 & 7.89 \\ 
		& vot & 82.84 & 83.55 & 92.53 & 86.31 & 5.64  & 4.26 & 2.90 & 4.27 \\ 
		\multirow{2}{*}{Supervised stage via uncertainty}   & avg & 85.79 & 85.75 & 92.11 & 87.88 & 6.66  & 6.60 & 4.59 & 5.95 \\ 
		& vot & 86.51 & 86.50 & 92.17 & 88.39 & 3.93  & 3.52 & 2.76 & 3.40 \\ 
		\multirow{2}{*}{Unsupervised stage via uncertainty} & avg & 84.55 & 84.73 & 92.26 & 87.18 & 7.42  & 5.75 & 4.50 & 5.89 \\ 
		& vot & 85.29 & 85.37 & 92.45 & 87.70 & 4.42  & 3.41 & 2.72 & 3.52 \\ \hline
	\end{tabular}
\end{table*}
\subsection{Implementation Details}
For the sake of speed and accuracy, we leverage the well-known lightweight E-net\cite{enet} as our basic segmentation network. This architecture is one of the most popular models for segmentation. To adapt the E-Net as a Bayesian network to estimate the uncertainty, one dropout layer with a dropout rate of $0.5$ is added between the encoder and decoder of the E-Net. All the datasets are applied random rotation, random crop, and flip as augmentation strategies. We set $\epsilon$ to 0.03 in FSGM and 10 in VAT.
The learning rate decreased with a factor of 10 every 90 epochs, and it is 10 every 100 epochs in SCGM experiments. To describe the level of supervision, we vary the ratio $l_a$ (0 $\leq$ $l_a$ $\leq$ 1) of labeled samples in our experiments. 

For all experiments, the setting of $\lambda$ is with a dynamic strategy. We used a ramp-up strategy followed the Gaussian ramp-up curve. And for all the hyperparameter $\lambda$ settings, we follow ~\cite{Deep-co-training}. The framework is implemented with Pytorch library, trained on one NVIDIA 2080Ti GPU.

\subsection{Comparison with semi-supervised methods}
Our uncertainty-aware co-training method is compared against several recent state-of-art semi-supervised methods in the medical domain:
\paragraph{\textbf{Mean Teacher}}: Mean Teacher (MT) is a method using multiple deep CNNs for semi-supervised segmentation. And it is an effective approach that averages model weights instead of predictions.
\paragraph{\textbf{Uncertainty-aware Mean Teacher}}: Uncertainty-aware Mean Teacher (UAMT) applies uncertainty estimation to MT. 
\paragraph{\textbf{Deep-Co-trianing}}:  Deep-Co-training (CT) shares information between simultaneously trained models, while preserving their diversity.

We also compare our method with the full supervised method of the same amount of labeled images as the semi-supervised method, referred to as "Part". "IND" means the performance of individually-trained models (Independent). We report the performance of a fully supervised baseline ("Full")  which means training a single model with all available datasets.

For these baselines, we follow the same learning rate decay, weight scheduler, data augmentation setting, and optimization as for our method. For MT and UAMT, we set EMA (Exponential Moving Average) parameters $\alpha$ to 0.99. We report both the average performance of individual models ("avg") and the performance of combining the predictions of all models using a voting strategy ("vot") called ensemble soft voting. This strategy leads to a higher accuracy than the prediction of individual models for both DSC and HD. In order to avoid contingency, we compute the average and standard deviation over three runs with different random seeds. 

All the compared baseline results reported in our paper are reimplemented by us.

\subsection{Experimental results}
\subsubsection{ACDC dataset}
Our uncertainty-aware method is first evaluated on the ACDC dataset with $l_a$ = 0.2. Tab.~2. presents the quantitative results. Compared with other methods, our approach achieves higher overall performance on the test dataset in terms of DSC and HD. Especially, our approach achieves improvement over co-training with an overall performance increase of 2.29$\%$ in DSC and 0.49$mm$ in HD, and the standard deviation of multiple trials (three runs) of the results of our approach is the smallest. With only 20$\%$ labeled images, we are only 1.8$\%$ in DSC and 0.39$mm$ in HD behind the 100$\%$ label ratio supervision. Some examples of results from the test dataset are shown in Fig.~5. We can see our method gives contours closer to ground truth (GT), with more accurate segmentation in details between different regions.

Also, we evaluate how label ratios impact results in a dual-view setting. Tab.~3. shows the results for different labeled data ratios: 5$\%$, 10$\%$, 20$\%$, 30$\%$, 40$\%$, 50$\%$. We can easily find that as the label ratio increase, mean DSC values increase sharply, and mean HD values decrease. In all cases, our approach leads to a better performance in DSC and HD than training models separately and deep co-training method. Our approach has more obvious advantages with a low label ratio. With 5$\%$ label ratio, our method outperforms Deep co-training 2.73$\%$ in DSC and 1.34$mm$ in HD.

\subsubsection{SCGM dataset}
We evaluate our method on the task of segmenting spinal cord grey matter in images from the SCGM dataset. The SCGM dataset is from four different clinical centers, so different parameters are applied in collecting the MRI images. We only used a few labeled images (i.e., only 30 images from one center), and test sets are from the other centers. The labeling ratio is about 6.5$\%$. Due to the different data sources, it is more difficult to extract the image features of the samples. This also leads to low segmentation accuracy of semi-supervised methods in this task.  The use of uncertainty solves this problem well, models can learn effective semantic features from different centers according to the uncertainty.

The results are summarized in Tab.~4. Our approach gives a mean DSC of 5.96$\%$ and HD of 6.35$mm$ better than the best baseline (deep co-training). And our method gives a mean DSC of 12.94$\%$ and HD of 23.26$mm$ better than the performance of the fully supervised baseline. Fig.~6. shows the segmentation results on the test dataset. In some difficult-to-recognize images, fully supervised training cannot even segment the lesion area, but our approach can complete the segmentation task to a certain extent.

\subsubsection{Spleen dataset}
We also validate the effectiveness of our uncertainty-aware deep co-training method on the task of segmenting spinal cord grey matter in images from the Spleen dataset. We repeated our experiments on the Spleen dataset consisting of 2D slices of CT scans resized to a resolution of 256 $\times$ 256 pixels. Tab.~5. summarizes the experimental results. We see that the Deep co-training method's performance is almost the same as the fully supervised baseline. In case of better stability (standard deviation) of the results, the accuracy has been improved and surpassed full supervision after taking uncertainty into concern. Our method improves accuracy over DCT-seg from 94.09$\%$ to 95.45$\%$ and HD from 6.86$mm$ to 5.57$mm$. Examples of segmentation results obtained by tested methods are given in Fig.~7. 

Semi-supervised learning pursues using less labeled images to achieve better segmentation accuracy. We show the performance of our approach, DCT-seg, and individually trained models(independent) on smaller label ratios: 5$\%$ and 10$\%$. Tab.6. gives the mean DSC with standard deviation. Our approach is still in the leading position among all semi-supervised methods. 

\subsubsection{Abalation analysis of our method}
We do ablation studies to prove that our method works in both supervised and unsupervised stages on the ACDC dataset. The results are shown on Tab.6.

\textbf{In supervised stage}: As mentioned before, the best time to introduce uncertainty to the full supervision phase is from the beginning. Uncertainty used here allows models in the network to obtain characteristics of the target area better under full supervision. We can see from Fig.~8. that the main effect of uncertainty in the full supervision stage is in the first 30 epochs. We can tell from the label that the uncertainty maps focus more on the lesion area. After adding uncertainty to the full supervision stage, the method can achieve improvement over co-training with an overall all performance increase of 2.23$\%$ in DSC and 0.57$mm$ in HD. 

\textbf{In unsupervised stage}: The best time at this stage is to leverage uncertainty after models have a certain understanding of the lesion area's semantic features. After experiments per 10 epochs from 0, it is finally determined that the uncertainty will be leveraged after the 20th epoch. So, we visualize the uncertainty map from 0 epochs to the last in Fig.~9. We can see that the uncertainty maps are increasingly focusing on the lesion areas. Obviously, uncertainty always plays a role throughout the experiment. As the training progresses, the color of the lesion areas of the uncertainty maps becomes lighter, which means the predictions of the two models for the same unlabeled image are gradually approaching, but there is still a certain difference. Only introduce uncertainty to the unsupervised stage, we can achieve an improvement over co-training with an overall all performance increase of 1.39$\%$ in DSC and 0.75$mm$ in HD.

\section{Conclusion}
\label{sec6}
In this paper, we propose a novel uncertainty-aware deep co-training method for three medical image segmentation tasks. Our approach uses uncertainty in both the supervised learning stage and the unsupervised learning stage in the deep co-training method. We use uncertainty obtained from Monte Carlo Sampling to guide the training process purposely. We validate our method on three challenging medical image datasets. The comparison with other semi-supervised methods confirms the effectiveness of our approach. Our uncertainty-aware co-training method achieves the performance 1.88$\%$ away from supervision with only 20$\%$ label on ACDC dataset, and an increase of 12.94$\%$ in terms of DSC over fully supervised method on SCGM dataset. And for the Spleen dataset, we get 0.89$\%$ more than the fully supervised approach in DSC. In future work, we will investigate the effect of different uncertainty estimation manners and apply our approach to other semi-supervised medical image segmentation tasks.

\bibliographystyle{elsarticle-num} 
\bibliography{batchCBC}





\end{document}